\documentclass[11pt,a4paper]{article}
\usepackage{times,latexsym}
\usepackage{url}
\usepackage[T1]{fontenc}
\usepackage{graphicx}
\usepackage{latexsym}
\graphicspath{{figures/}}
\usepackage{url}
\usepackage[]{algorithm2e}
\usepackage{fixltx2e}
\usepackage{booktabs}

\usepackage[acceptedWithA]{tacl2018v2}

\usepackage{xspace,mfirstuc,tabulary}

\newif\iftaclinstructions
\taclinstructionsfalse 
\iftaclinstructions
\renewcommand{\confidential}{}
\renewcommand{\anonsubtext}{(No author info supplied here, for consistency with
TACL-submission anonymization requirements)}
\newcommand{\instr}
\fi

\iftaclpubformat 

\else

\fi

\title{Conversation Graph: Data Augmentation, Training and Evaluation for Non-Deterministic Dialogue Management}

\author{
 Milan Gritta, Gerasimos Lampouras, Ignacio Iacobacci \\
 Huawei Noah's Ark Lab, London, United Kingdom \\
  {\sf milan.gritta, gerasimos.lampouras, ignacio.iacobacci@huawei.com} \\
}

\date{}

\begin{document}
\maketitle
\begin{abstract}

Task-oriented dialogue systems typically rely on large amounts of high-quality training data or require complex handcrafted rules. However, existing datasets are often limited in size considering the complexity of the dialogues. Additionally, conventional training signal inference is not suitable for non-deterministic agent behaviour, i.e. considering multiple actions as valid in identical dialogue states. We propose the Conversation Graph (ConvGraph), a graph-based representation of dialogues that can be exploited for data augmentation, multi-reference training and evaluation of non-deterministic agents. ConvGraph generates novel dialogue paths to augment data volume and diversity. Intrinsic and extrinsic evaluation across three datasets shows that data augmentation and/or multi-reference training with ConvGraph can improve dialogue success rates by up to 6.4\%.

\end{abstract}

\section{Introduction}
\label{sec:intro}

Dialogue systems research focuses on the natural language interaction between a user and an artificial conversational agent. Current trends lean towards end-to-end models \cite{bordes2016learning,miller2017parlai} while modular systems tend to be the preferred approach in industrial applications \cite{bocklisch2017rasa,burtsev2018deeppavlov}. At the core of a modular conversational agent is dialogue management (DM) whose function is to exchange information with a user, update the agent's internal state and plan its next action according to a policy
\cite{young2000probabilistic}. The dialogue manager then collects the user's response and the process repeats until the conversation ends. More specifically, task-oriented dialogue managers are aiming towards the completion of some specific task or tasks (e.g. booking services or buying products).


Machine-learned policies for DM require large amounts of high-quality data to generalise to a variety of conversational scenarios \cite{asri2017frames,shah2018building,budzianowski2018multiwoz, rastogi2019towards}. However, given the complexity of some tasks, the datasets are often limited in size. Reinforcement learning approaches  \cite{henderson2008hybrid,bordes2016learning,miller2017parlai,li2017end,gordon2020show} can replace the need for explicit training data by exploiting a custom-designed environment to infer the training signal for the policy. Unfortunately, such custom-designed environments may not be representative of how a user would interact with a conversational agent and their manual development is time-consuming and domain specific. Furthermore, modern conversational agents exhibit non-deterministic behaviour, i.e. they are able to take different but equally valid actions in identical dialogue states. Conventional agent training and evaluation do not support the non-deterministic nature of conversational datasets as only a single fixed target is considered per inference, penalising valid model predictions. \\

To address these problems, we propose the Conversation Graph (ConvGraph), a DM framework for data augmentation, which also enables the training and evaluation of non-deterministic policies. If we consider each dialogue as a sequence of dialogue states, alternating between the agent and the user, we assemble a graph structure by unifying matching states across all conversations in the dataset. New dialogue paths can then be traversed in the graph resulting in novel training instances that can improve the DM policy. The unification simultaneously collects all valid actions at each point in the dialogue to facilitate the training and evaluation of non-deterministic policies.

We consider the contribution of this paper to be three-fold. First, we explore several augmentation baselines as well as variants of ConvGraph augmentation to show their impact on policy through intrinsic and extrinsic evaluation. We show that augmentation with ConvGraph leads to improvements of up to 6.4\% when applied in an end-to-end dialogue system. In addition, we propose a loss function that takes advantage of ConvGraph's unified dialogue states to increase the success rate by up to 2.6\%. Finally, we exploit ConvGraph to introduce a multi-reference evaluation metric for non-deterministic dialogue management.

\section{Background}
\label{background}

The idea of using graphs in dialogue systems is not new \cite{larsen1994rapid,schlungbaum1996dialogue,agarwal1997towards}, but it is limited to graphs representing flow charts where each node is an action step in a sequence, without a specific semantic importance assigned to the nodes themselves~\cite{aust1995dialogue,warnestaal2005modeling}. On the other hand, the ConvGraph dialogue state information is encoded in a structured way for each node, allowing the unification of nodes \textit{across} conversations. It is primarily this structured representation of dialogues that enables the use of the graph for data augmentation by traversal of new dialogue paths. It also allows for non-deterministic training and evaluation by referencing the graph to validate model predictions.

\subsection{Data Augmentation}

Data augmentation for dialogue management is relatively unexplored and limited to increasing training data volume with (random) data duplication/recombination \cite{bocklisch2017rasa} or general machine learning data transformations such as oversampling and downsampling of existing samples \cite{chawla2002smote}. We explore variants of both strategies as additional baselines. Related to our work is a recent paper on Multi-Action Data Augmentation~\cite[MADA]{Zhang2019TaskOrientedDS}. Similarly to us, they are leveraging the fact that a non-deterministic agent can take different actions given the same dialogue state. However, our methodologies then diverge sharply. While we pursue data augmentation for DM, MADA is applied towards the task of context-to-text Natural Language Generation \cite{wen2016network}.
This necessitates fundamental differences such as MADA not abstracting away slot values because its states must be unified on literal values and database results to generate the required natural language response. MADA additionally does not consider any dialogue history hence this approach is not suitable for dialogue management data augmentation. 

\subsection{Training and Evaluation}

ConvGraph enables the training and evaluation of non-deterministic agents. When multiple actions are equally valid in a particular dialogue state, conventional 'pointwise' machine learning training and evaluation adversely influences, even penalises otherwise correct predictions. Multi-reference training (sometimes called "soft loss") has been used for this reason, e.g. to improve the application of Maximal Discrepancy to Support Vector Machines \cite{anguita2011maximal} as well as to boost performance for noisy image tag alignments \cite{liu2012noisy}. Multi-reference evaluation has been the standard for language generation tasks in NLP, as it accounts for the non-deterministic nature of language output. In some cases, both soft training and evaluation have been used, e.g. for decision tree learning with uncertain clinical measurements \cite{nunes2020decision} to mitigate the impact of hard thresholds.

\begin{figure*}[]
\includegraphics[width=0.70\textwidth]{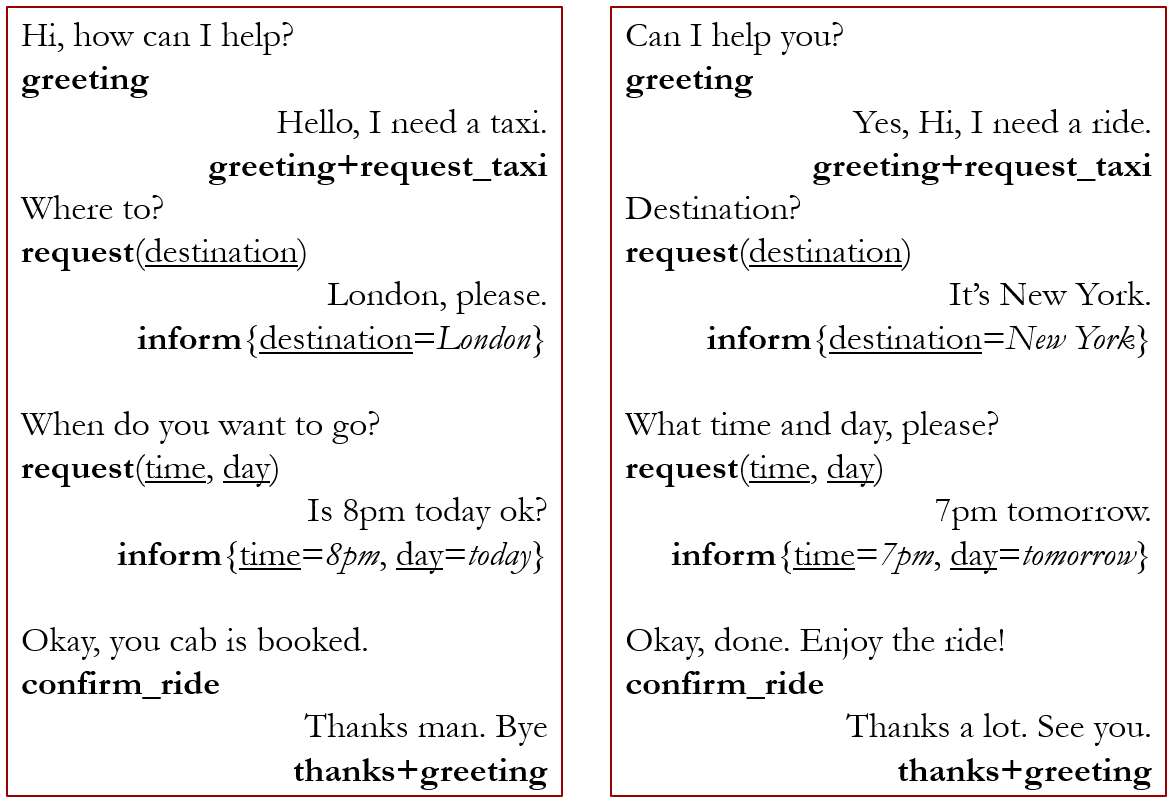}
\centering
\caption{Two dialogues with different utterances (in plain text) but identical \textit{da} sequences of (multi) \textbf{intents} and \underline{slots}, entities (in plain text) and \textit{slot values}. These dialogues follow \textbf{identical edges/nodes} in ConvGraph.}
\label{dialogues_fig}
\end{figure*}

\section{The Conversation Graph}
\label{CG}

This section describes how dialogues are unified into a graph representation useful for data augmentation, model training and evaluation.

\subsection{Key Concepts}

In a task-oriented dialogue system, the user interacts with an agent through natural language in order to achieve a specific goal. Every user or agent utterance corresponds to a \textbf{dialogue act} ($da$) (see Figure \ref{dialogues_fig}), i.e. an abstract representation typically consisting of an intent and a set of slots with corresponding values. The \textbf{intent} denotes the abstract communication goal of the user or agent and the \textbf{slot} ($s$) - \textbf{value} ($v$) pairs encode the entities provided in an utterance. For example, $inform\{destination=London\}$ denotes that the communication goal is to inform the listener that the value of the destination slot is 'London'. The values of all slots in a given turn constitute the \textbf{belief state} ($bs$) such that $bs$=[(\textit{s}\textsubscript{1}=\textit{v}\textsubscript{1}), (\textit{s}\textsubscript{2}=\textit{v}\textsubscript{2}),\ ...\  (\textit{s}\textsubscript{n}=\textit{v}\textsubscript{n})]. 
After each dialogue turn, the $bs$ is updated as the dialogue proceeds towards the goal. Note that we may observe multiple intents at each turn, depending on the design of the dialogue system. We finally define the \textbf{dialogue state} ($ds$), which is a concatenation of $bs$ and $da$ at each turn.
\subsection{Construction}

\begin{figure*}[t]
\includegraphics[width=1.\textwidth]{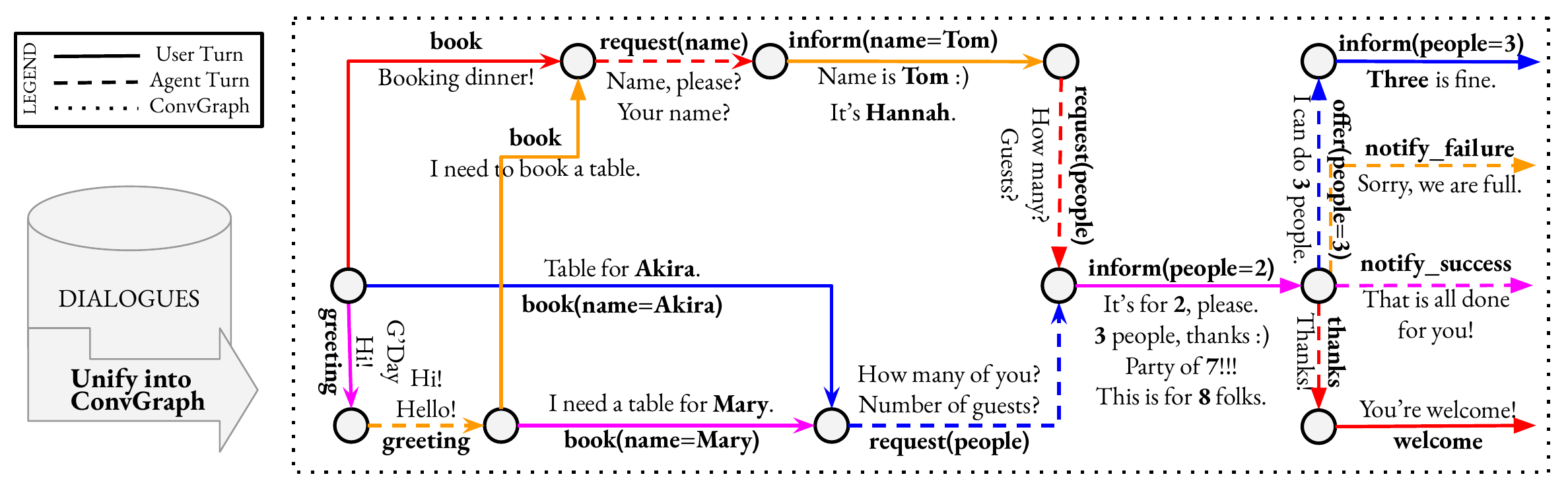}
\centering
\caption{ConvGraph unification with four fictitious dialogues, which intersect on identical dialogue states.}
\label{cf_fig}
\end{figure*}

We treat each dialogue \emph{D} as a sequence of \emph{n} encoded turns such that $D$=$[ds\textsubscript{0}, ds\textsubscript{1}, ds\textsubscript{2}, ..., ds\textsubscript{n}]$ where $ds\textsubscript{0}$ is the start state and $ds\textsubscript{n}$ is the end state. ConvGraph is defined as a directed graph $ConvGraph$=$(N, E)$ where $N$ is a set of nodes, each corresponding to a dialogue state $ds_{i}$ and $E$, which is the set of edges (transitions) between any two nodes. An edge corresponds to a user or an agent dialogue act. Its frequency is also recorded, as observed in the data. Algorithm~\ref{cg_algorithm} shows how multiple dialogues (\textit{DS}) are converted into a conversation graph such that nodes that are identical are unified. As a result, dialogue sequences intersect on common nodes (see Figure~\ref{cf_fig}). During this unification, we infer which actions are valid, given the same $ds$. We additionally append an artificial final state to each dialogue to explicitly mark the end of the conversation for datasets where a "task\_complete" indicator is not present \cite{shah2018building}.

\begin{algorithm}[t]
\SetAlgoLined

\KwData{Dialogues $DS$}
\KwResult{ConvGraph $CG$}
 $ds_{0}$ = zeros($|ds|$)  //empty start state
 
 $CG$ = $empty\ directed\ graph$
 
 \For{$D$ in $DS$}{
 
  $lastState$ = $ds_{0}$
  
  \For{$turn$ in $D$}{
  
  $ds$ = encode($turn$[$bs$], $turn$[$da$])

  $CG$.newEdge($lastState$, $ds$)

    $lastState$ = $ds$
  }
}
\caption{Dialogues unified into a graph.}
\label{cg_algorithm}
\end{algorithm}

\subsection{Data Augmentation}
\label{data_augmentation}

The aim of DM in a modular dialogue system is to learn a policy to predict an appropriate agent action (dialogue act $da$) at turn $t$, conditioned on the history of previous dialogue states $PDS$. We define a policy $\pi_{\theta}(da_{t}|PDS)$ where $PDS$=[$ds_{t-1}$, $ds_{t-2}$, .., $ds_{t-n}$] and $n$=history length. Machine-learned policies can benefit from additional training instances in order to better generalise and reduce overfitting. This leads us to ConvGraph's first application, the inference of additional training signal that can be used to train a DM policy.

\paragraph{Augmentation by Most Frequent Sampling} is our main method. In order to generate training instances from ConvGraph, we visit all nodes and extract ($da_{t}$ | $PDS$) pairs for the policy. Since an exhaustive traversal of ConvGraph is unrealistic due to its size and connectivity on the datasets we considered, we need a strategy to select the most useful pairs. In preliminary experiments, we performed uniform sampling amongst the outgoing edges at each agent node. This was not promising as it ignored the likelihood of agent actions. Instead, for each agent node in ConvGraph, we exclusively choose the most frequent outgoing edge, as observed in the original data. This process pairs frequent actions with a new history (context) thus creating new training examples. It also results in a reduction of actions at each agent node, decreasing conflicting training signals for the policy. This approach for inferring training instances, which can be combined with the original data, resulted in the most effective experimental setup (Results \ref{results}). We refer to this method as MFS henceforth. 

\paragraph{Oracle Augmentation} is featured in our experiments to explore the performance impact of oracle-guided augmentation. An oracle in this context represents an information source that can be queried to obtain additional (but \textit{incomplete}) information about the development and test set. More specifically, while generating novel instances with ConvGraph, the oracle can tell us which novel training examples occur in the development and/or test set. The oracle confirms whether a new training example will be informative to the DM policy and likely lead to higher development/test scores. Adding these instances to the original training data creates another challenging baseline policy. Please note that this baseline is designed for theoretical comparisons only as obtaining this type of strong information source is not possible in real settings.

\subsection{Additional Augmentation Approaches}
\label{other_baselines}

To the best of our knowledge, there are few comparable data augmentation methods for DM. The following are the most relevant baselines that were included in our experiments.

\paragraph{Downsampling} removes \textit{duplicate} training examples thus results in only \textit{unique} instances being included in training. This action therefore reduces the size of (and balances) the train set. Downsampling is related to SMOTe, the Synthetic Minority Oversampling Technique \cite{chawla2002smote}, particularly its later variants \cite{han2005borderline}, which aim to balance the training data by oversampling rare cases. This has been shown to improve classifier performance \cite{maciejewski2011local,ramentol2012smote}. 

\paragraph{Data Duplication} is an adaptation of dialogue concatenation and/or recombination, available in some dialogue systems although this has not been rigorously evaluated \cite{bocklisch2017rasa}. Our implementation takes care to combine the state sequences in a way that does not introduce inconsistent state transitions, e.g. starting a new conversation without resetting the dialogue state, which would result in skipping some required steps.

\subsection{Multi-Reference Evaluation}
\label{softeval}

Before we describe the full evaluation procedure, we briefly introduce the Evaluation Graph (denoted 'eval' in Table \ref{data_stats}). This is a regular conversation graph constructed from \textit{all data splits} using Algorithm \ref{cg_algorithm}. The EvalGraph serves as a reference tool used to look up the list of valid actions for each agent node.\\

The EvalGraph allows us to pool all observed agent actions into a single graph and use it to score policy predictions. We evaluate the predicted dialogue act $\hat{y}$ against all of the valid target dialogue acts $Y$ and report the greatest score (see Equation \ref{s_fscore}). For example, if $Y$ = [\textit{request(time, date)}, \textit{request(time)} and \textit{request(date)}] and $\hat{y}$ = \textit{request(time)}, then the maximum score is awarded. In our experiments, this modification of the F-Score \cite{JMLR:v12:pedregosa11a} is referred to as \textit{SoftF1}. The running time is approximately an order of magnitude slower.

\begin{equation}
\small
SoftF1(\hat{y},Y)=max([F1(y,\hat{y})\ \forall y \in Y])\label{s_fscore}
\end{equation}

\subsection{Multi-Reference Training}
\label{softloss}

Most conversational datasets, including ones featured in the evaluation (see \ref{datasets}), contain user interactions with non-deterministic agents. Given more than one valid response in a given state, conventional single-reference 'hard' training penalises the model for making a valid prediction. Propagating such a loss is likely to lead to a deteriorating dialogue management policy.

\begin{equation}
\footnotesize
    BCE(y,\hat{y})=-\sum_{i=1}^{|y|} y_{i}\log(\hat{y}_{i}) + (1 - y_{i})\log(1 - \hat{y}_{i})\label{bce}
\end{equation}

We therefore modify the Binary Cross-Entropy (BCE) loss from Equation \ref{bce} to propose the Soft Binary Cross-Entropy (SBCE) seen in Equation \ref{sbce}. SBCE uses ConvGraph to compute losses for all valid actions $Y$ in a given dialogue state $ds$. SBCE then propagates the lowest loss for $\hat{y}$. Training time with SBCE increases by an order of magnitude since multiple references are considered.

\begin{equation}
\small
    SBCE(\hat{y},Y)=min([BCE(y,\hat{y}) \ \forall y \in Y])\label{sbce}
\end{equation}

\section{Experimental Setup}
\label{eval}

Next, we describe the platform, datasets and metrics used throughout our experiments.

\subsection{Datasets}
\label{datasets}

There are two main approaches to dataset construction for dialogue systems. Machines talking to machines (M2M) is a data generation framework that makes use of rule-based user and agent simulators that interact to generate sequences of dialogue acts. Also known as Dialogue Self-Play \cite{shah2018building}, crowd workers proceed to lexicalise them to produce corresponding natural utterances. This approach has two main limitations: (i) both simulators must be hand-coded and (ii) there is no guarantee that the simulators generate realistic conversations. Other examples of such datasets include AirDialogue \cite{wei2018airdialogue} and Schema Guided Dialogue \cite{rastogi2019towards}. The second approach is collecting dialogues through a Wizard of OZ~\cite[WOZ]{dahlback1993wizard,strausswizard2006} setting, which has been used to gather the DSTC2 \cite{henderson2014second}, WOZ 2.0 \cite{wen2016network}, Frames \cite{asri2017frames}, Microsoft E2E Challenge \cite{li2018microsoft} and MultiWOZ \cite{budzianowski2018multiwoz} datasets. The collection process involves two humans conversing, one acting as the agent and the other as the user. In standard Wizard of OZ, the user is led to believe that the agent is artificial rather than a human. This helps ensure that gathered dialogues reflect how users interact with machine-driven agents.\\

For our experiments, we use three datasets with original splits, both human and machine generated to ensure the applicability of our methods to a variety of conversational scenarios. These are the \textbf{movie} and \textbf{restaurant} partitions of M2M and the extended version of MultiWOZ 2.0 \cite{lee-etal-2019-convlab} with user intent annotations. Note that the original dataset has since been corrected and released as MultiWOZ 2.1 \cite{eric2020multiwoz}. The descriptive statistics for each dataset can be found in Table \ref{data_stats}. ConvGraph requires user/system intents and other dialogue annotations thus several aforementioned datasets were not compatible.

\subsection{Test Set Deduplication}

All test sets contain some degree of duplicate instances. Table \ref{data_stats} shows the highest duplication (\# unique / \# instances) for M2M restaurant ($\sim$61\% unique), followed by M2M movie ($\sim$75\% unique), and MultiWOZ ($\sim$96\% unique). For a more complete evaluation, we present results on two test sets for each dataset. Besides the original data, we also evaluate on a deduplicated test set. This ensures that changes in model performance are not disproportionately influenced by duplicated instances.

\subsection{End-to-end User Simulation}
\label{convlab_desc}

ConvLab \cite{lee-etal-2019-convlab} is an end-to-end dialogue system platform built to support the Multi-Domain Wizard-of-Oz dataset~\cite[MultiWOZ]{budzianowski2018multiwoz}, a collection of human-to-human conversations spanning multiple domains and one of the largest annotated task-oriented corpora for dialogue. ConvLab allows an agent to interact with the user simulator via dialogue acts, and supports reinforcement learning, supervised learning and rule-based agents. It can be used as an evaluation platform to test modular, task-oriented conversational agents in an end-to-end fashion and has been used in this capacity at the \textit{Eighth Dialog System Technology Challenge}~\cite{kim2019eighth}. We use ConvLab for extrinsic evaluation of DM policies. Note that the platform does not support the entirety of actions occurring in MultiWOZ. We adapt our policies' output action space accordingly, so that we are compatible with ConvLab and able to duly perform the extrinsic evaluation.\footnote{Please note that we are \textit{not} excluding any portion of the MultiWOZ data. When multiple actions are taken in the same turn, ConvLab handles them as a single concatenated action, e.g. inform(departure) and inform(destination) are treated as inform(departure+destination). Due to this paradigm, every action combination needs to be treated distinctly by ConvLab. As the number of combinations is large, the user simulator is restricted to the 300 most frequent combined actions. Any policy's output needs to comply to the restricted action space.}

\subsection{Metrics}
\label{metrics}

We evaluate policies intrinsically and extrinsically, the latter through the ConvLab user simulator and limited to MultiWOZ since the dialogue system framework for M2M is proprietary and unavailable. For extrinsic evaluation, the main automatic metric is success rate \cite{kim2019eighth}, averaged over 1000 conversations or episodes. A dialogue is considered successful if all informable slots (what the agent needs to complete the task) and requestable slots (what the user wants to know) were correctly filled. We also report the average number of dialogue turns. For intrinsic evaluation, we use SoftF1 scores (Section~\ref{softeval}) alongside the conventional F-Scores \cite{JMLR:v12:pedregosa11a}. We refer to the latter as \textit{HardF1} as it is computed strictly against a single target $y \in Y$, even if another prediction $\hat{y}$ was valid under a 'soft' evaluation because $\hat{y} \in Y$. In other words, when the agent is able to predict multiple different valid actions $Y$, the SoftF1 score reports the lowest error for \textbf{any} $y \in Y$ while HardF1 reports the error for \textbf{exactly one} $y \in Y$. Using only HardF1 can lead to unfair penalties for otherwise correct predictions. HardF1 and SoftF1 are computed using 'samples' averaging, that is, score each prediction \textit{separately}, then average for an overall score. This is the most realistic scenario for the DM task as it evaluates the quality of \textit{each interaction} without pooling or averaging predictions over multiple turns. 'Micro' and 'macro' averages can underestimate or exaggerate changes in predictions, particularly for infrequent classes.

\begin{table*}[]
\begin{center}
\begin{tabular}{l|c|c|c|c|c|c|c}
\toprule \textbf{Dataset} & \textbf{Edges} & \textbf{Repetition} & \textbf{Nodes} & \textbf{MND} & \textbf{\# dial} & \textbf{\# instances} & \textbf{\# unique}\\ \midrule
M2M-R train & 1,767 & 84.3\% & 725 & 2.48 & 1,116 & 5,059 & 2,492\\
M2M-M train & 978 & 72.5\% & 472 & 2.01 & 384 & 1,589 & 1,063\\
M-WOZ train & 82,097 & 27.7\% & 55,783 & 1.41 & 8,434 & 56,753 & 51,288\\ \midrule
M2M-R val & 935 & 64.4\% & 533 & 1.8 & 349 & 1,140 & 812\\
M2M-M val & 498 & 56.1\% & 279 & 1.86 & 120 & 507 & 397\\
M-WOZ val & 12,744 & 13.5\% & 10,140 & 1.27 & 999 & 7,365 & 7,116\\ \midrule
M2M-R test & 1,541 & 74.7\% & 706 & 2.26 & 775 & 2,661 & 1,631\\
M2M-M test & 842 & 63.8\% & 436 & 1.82 & 264 & 1,100 & 822\\
M-WOZ test & 12,659 & 14.1\% & 10,094 & 1.27 & 1,000 & 7,372 & 7,076\\ \midrule \midrule
M2M-R eval & 2,563 & 87.2\% & 959 & 2.69 & 2,240 & 8,860 & 3,795\\
M2M-M eval & 1,458 & 79.6\% & 619 & 2.24 & 768 & 3,196 & 1.850\\
M-WOZ eval & 101,864 & 28.8\% & 67,517 & 1.44 & 10,433 & 71,487 & 64,054\\
\bottomrule
\end{tabular}
\end{center}
\caption{Statistics of standard data splits; \textit{eval} = train + val + test splits; \textit{MND} = Mean Node Degree, the average number of outgoing edges (higher number means a denser graph); \textit{\# instances} = number of training instances of which \textit{\# unique} are unique; \textit{\# dial} = number of dialogues; \textit{Repetition} = percentage of edges visited more than once.}
\label{data_stats}
\end{table*}

\subsection{Policy Implementation}
\label{experimental}

As the approaches we examine are \textit{orthogonal} to the policy implementation, in order to minimise the influence of hyperparameter/architecture choice, we have fixed the model for training and evaluation across all experiments. We have used an LSTM~\cite{hochreiter1997long} model with default PyTorch hyperparameters to train all DM policies. We learn a policy $\pi_{\theta}(da_{t}|PDS)$ where $t$ is the current agent turn, $PDS$ = [$ds_{t-1}$, $ds_{t-2}$, .., $ds_{t-n}$] are the previous dialogue states and $n$ is the history length. In our experiments, we set $n$ to 3, 4 and 5. For brevity, the Results (section \ref{results}) feature scores with history set to 4 as the differences are negligible. We proceed to encode $PDS$ with the LSTM with a hidden layer size of $256$. We explored several hidden sizes between 64 (underfitting) and 512 (overfitting) but the relative rank and differences between experimental setups were not affected. The LSTM output is passed through a ReLu \cite{nair2010rectified} activation followed by a linear layer with a sigmoid activation and size equal to the output size. The output size is the number of distinct labels in the observed dialogue acts. The input size is equal to $|ds|$, i.e. output size + the encoded belief state size. The model parameter counts are as follows: M2M movie (299K) with input size of 31 and output size of 12, M2M restaurant (314K) with input size of 45 and output size of 16 and MultiWOZ (707K) with input size of 355 and output size of 309. Training with batch size of 32, the development set HardF1 score was monitored for early stopping with a patience of 5 for the M2M models and 3 for MultiWOZ policies. For an illustration of computational requirements, a MultiWOZ experiment with SBCE loss training on an NVIDIA GeForce RTX 2080 GPU Ti 11GB takes $\sim$2.5 hours.\\

The output space of dialogue management models can be framed as either \textit{multi-class classification} or \textit{multi-label classification}. In some dialogue systems \cite{bocklisch2017rasa}, actions are predicted and executed one at a time, which lends itself to multi-class classification with a Cross-Entropy loss as the probability of the target label is maximised while all other label probabilities are minimised. However, in most conversational research datasets (\ref{datasets}), several target labels are jointly predicted. We consider multi-label classification with BCE (Equation \ref{bce}) or SBCE loss function (Equation \ref{sbce}) to be more suitable for this type of DM task. An increase in the number of dialogue acts means that the output vector size would grow at a \textit{constant} rate when considering multi-label classification but would grow \textit{exponentially} with multi-class classification. This is not scalable beyond but the simplest dialogue systems. Therefore, multi-label classification allows for a highly expressive agent using a small target vector while also being more sample-efficient.

According to extrinsic evaluation in ConvLab, this configuration leads to a 73.4\% success rate and 10.11 turns as the average conversation length (Table \ref{convlab_table}). 
Therefore, our (multi-label) baseline achieves stronger results than the baseline used in the \textit{Eighth Dialog System Technology Challenge}~\cite{li2020results}, which reached a 61\% Success Rate and 11.67 turns on average. Note that our data-augmentation approach and loss function are \textit{orthogonal} to the choice of the DM model.

\subsection{Statistical Significance}
Ten models were trained for each experiment to determine the mean and variance under various random seeds. We then perform an Analysis of Variance followed by a two-tailed t-test (samples with unequal variance). In Tables \ref{hard_loss_table} and \ref{soft_loss_table}, significant differences are noted with an asterisk (*). 

\begin{table*}[t]
\centering
\setlength{\tabcolsep}{4.5pt}
\renewcommand{\arraystretch}{1.2}
\small
\begin{tabular}{@{}l||l|l||l|l||l|l||l|l||l|l||l|l@{}}
\toprule
\multicolumn{1}{c||}{Datasets} & \multicolumn{2}{c||}{BASE (B)} & \multicolumn{2}{c||}{D-SAMPLE} & \multicolumn{2}{c||}{DATA DUPL} & \multicolumn{2}{c||}{MFS} & \multicolumn{2}{c||}{MFS + B} & \multicolumn{2}{c}{ORACLE + B} \\ \midrule
Metrics & H-F1 & S-F1 & H-F1 & S-F1 & H-F1 & S-F1 & H-F1 & S-F1 & H-F1 & S-F1 & H-F1 & S-F1\\\midrule
M2M (R) & 66.9 & 93.3 & 61.9* & 85.6* & 65.4* & 92.5 & 65.0* & 94.9 & 67.1 & \textbf{95.6*} & \textbf{68.8*} & 94.4 \\
M2M (M) & 65.4 & 90.7 & 65.4 & 91.5 & 64.9 & 91.4 & 65.4 & \textbf{95.2*} & 66.5* & 93.7* & \textbf{68.3*} & 93.1* \\
M-WOZ & \textbf{46.6} & 67.9 & 45.6* & 66.7* & \textbf{46.6} & 67.9 & 46.4 & \textbf{73.9*} & \textbf{46.6} & 69.0* & \textbf{46.6} & 67.5 \\ \midrule
M2M (R) & 61.7 & 91.8 & 59.4* & 86.6* & 60.8 & 91.0 & 60.0* & \textbf{95.8*} & 61.7 & 95.0* & \textbf{63.8*} & 92.5 \\
M2M (M) & 69.8 & 91.0 & 69.9 & 90.8 & 69.7 & 91.1 & 69.4 & \textbf{94.7*} & 69.9 & 93.1* & \textbf{72.2*} & 92.3* \\
M-WOZ & 45.9 & 66.5 & 45.3 & 65.6 & 45.8 & 66.4 & 45.5 & \textbf{72.6*} & 45.9 & 67.9 & \textbf{46.0} & 66.4 \\\midrule
M2M (R) & 65.8 & 93.4 & 60.1* & 85.8* & 64.7 & 92.6 & 64.8 & 95.9* & 66.3 & \textbf{96.0*} & \textbf{67.8*} & 94.1 \\
M2M (M) & 71.5 & 92.0 & 71.4 & 92.2 & 71.4 & 92.5 & 71.2 & \textbf{95.6*} & 71.7 & 94.1* & \textbf{73.8*} & 93.6* \\
M-WOZ & 46.5 & 67.1 & 45.4* & 65.8* & 46.4 & 67.2 & 46.3 & \textbf{73.7*} & 46.6 & 68.6* & \textbf{46.7} & 67.1 \\ \bottomrule
\end{tabular}
\caption{DM policies using the \textbf{\textit{BCE loss}}. Development Set (top three rows), De-duplicated Test Set (middle three rows), Original Test Set (bottom three rows). \textbf{*} denotes statistically significant (p < 0.05) from baseline. Bold figures indicate the best performing method(s) for each of the two metrics and for each row (experimental setting).}
\label{hard_loss_table}
\end{table*}

\section{Results and Analysis}
\label{results}

Tables~\ref{hard_loss_table} and \ref{soft_loss_table} show intrinsic evaluation results for DM policies with a history of 4, trained with BCE and SBCE loss, respectively. The H-F1 columns denote HardF1 scores, and the S-F1 columns denote SoftF1 scores. Table~\ref{convlab_table} presents the results of extrinsic evaluation in ConvLab.

\subsection{Graph (Dataset) Properties}
\label{density}

Conversational datasets have very distinct properties that will help us interpret the observed results. An intrinsic view of the data shown in Table \ref{data_stats} can be used to infer the approximate performance of the DM policies, even before any training.

\paragraph{The MultiWOZ graph} has the most complex dialogue state due to its multi-domain nature. Its EvalGraph, initiated from all MultiWOZ partitions, has $\sim$100K edges, which is 40 times more than M2M restaurant and almost 70 times more than M2M movie. However, the number of shared edges between the train graph (n=82K) and the development graph (n=12.7K) is only 2.7K. The train and test (n=12.7K) graphs also share 2.7K edges. Once featurised into training instances, the overlaps are even smaller, approximately 800 out of 7.3K for both test and dev sets. This enables us to predict that the available data is almost certainly insufficient to learn a supervised DM policy with F-Scores approaching 1.0, regardless of the model architecture. The dialogue state and target vector are too complex for the amount of data provided (71.5K instances of which $\sim$90\% is unique). Scenarios significantly different from ones observed in training effectively demand a zero-shot transfer to unseen test instances. However, since dialogue acts consist of multiple labels, the model is able to predict many of them correctly, which is why in extrinsic evaluation, the best policy successfully handles almost 80\% of dialogues.

\paragraph{The M2M Restaurant graph} is quite different from MultiWOZ beyond just the size difference. It is important to look at graph connectivity such as the average outgoing edges (MND in Table \ref{data_stats}) and the amount of repetition (percentage of edges visited more than once). This dataset has the highest density (2.69 MND) and repetition (87.2\%). It also has the most shared edges between train (n=1,767), development (n=935) and test (n=1,541) graphs. The train graph shares 629 and 951 edges with the development and test graphs respectively, a high percentage and the opposite of MultiWOZ. Perhaps unsurprisingly, the policy performance for both M2M restaurant and M2M movie is much higher than MultiWOZ. 

\paragraph{The M2M Movie graph} is approximately half the size of M2M restaurant in terms of nodes and edges. This means a lower dialogue state variety but it comes with a lower number of total dialogues (n=768 versus n=2,240). Repetition (79.6\% versus 87.2\%) is also lower while the shared edges between train (n=978), validation (n=498) and test (n=842) graphs are similar to M2M restaurant (65\% with dev and 59\% test). It is perhaps unsurprising that these datasets were generated with the same probabilistic automata, given their similarities. The repetition (or lack thereof) in dialogues strongly contributes to the differences across experimental results.

\subsection{SoftF1 and HardF1 Score}

Using two evaluation metrics helps us understand experimental results from different angles. The most important contribution of the addition of the SoftF1 score is being able to measure the error of the best valid response in each agent state, making evaluation fairer by not penalising otherwise correct predictions. While we recommend to strive to improve both scores, we observe that an improved SoftF1 score is more likely to lead to successful conversations as it promotes the choice of actions that lead to fewer policy errors in training. In extrinsic evaluation, this translates into a higher probability of the agent navigating a conversation from start to end hence higher SoftF1 scores with samples averaging are preferred for DM. 

\begin{table*}[t]
\centering
\setlength{\tabcolsep}{5.5pt}
\renewcommand{\arraystretch}{1.2}
\small
\begin{tabular}{@{}l||l|l||l|l||l|l||l|l||l|l||l|l@{}}
\toprule
\multicolumn{1}{c||}{Datasets} & \multicolumn{2}{c||}{BASE (B)} & \multicolumn{2}{c||}{D-SAMPLE} & \multicolumn{2}{c||}{DATA DUPL} & \multicolumn{2}{c||}{MFS} & \multicolumn{2}{c||}{MFS + B} & \multicolumn{2}{c}{ORACLE + B} \\ \midrule
Metrics & H-F1 & S-F1 & H-F1 & S-F1 & H-F1 & S-F1 & H-F1 & S-F1 & H-F1 & S-F1 & H-F1 & S-F1\\\midrule
M2M (R) & 54.5 & 98.0 & 55.2 & 97.8 & 55.2 & 97.3 & 54.3 & 97.0 & 54.7 & 98.0 & \textbf{55.5} & \textbf{98.6} \\
M2M (M) & 56.8 & \textbf{97.1} & 57.8 & 96.4 & 56.1 & 96.4 & \textbf{59.8*} & \textbf{97.1} & 56.9 & 96.2 & 56.2 & 96.6 \\
M-WOZ & 43.7 & 77.7 & 43.6 & 77.8 & 43.3 & 77.4 & \textbf{47.4*} & \textbf{78.2} & 43.2 & 77.5 & 43.6 & \textbf{78.2} \\ \midrule
M2M (R) & 48.8 & 96.7 & 49.5 & 96.9 & 48.9 & 96.1 & 48.1 & 96.6 & 49.0 & 96.8 & \textbf{49.7} & \textbf{97.9*} \\
M2M (M) & 61.1 & 96.4 & 61.3 & 95.7 & 60.6 & 95.7 & \textbf{61.4} & \textbf{96.6} & 60.9 & 95.7 & 60.5 & 95.8 \\
M-WOZ & 44.1 & 76.4 & 44.1 & 76.8 & 44.2 & 76.6 & \textbf{47.2*} & \textbf{77.0} & 43.2 & 76.0 & 43.9 & \textbf{77.0} \\\midrule
M2M (R) & 53.6 & 97.8 & 54.0 & 97.9 & 53.7 & 97.3 & 53.2 & 97.3 & 53.4 & 97.9 & \textbf{54.2} & \textbf{98.7*} \\
M2M (M) & 60.0 & 97.0 & 61.0 & 96.4 & 59.5 & 96.3 & \textbf{62.2*} & \textbf{97.1} & 60.0 & 96.3 & 59.5 & 96.5 \\
M-WOZ & 43.7 & 77.4 & 43.6 & \textbf{77.8} & 43.6 & 77.6 & \textbf{47.5*} & \textbf{77.8} & 42.9 & 77.0 & 43.4 & \textbf{77.8} \\ \bottomrule
\end{tabular}
\caption{DM policies using the \textbf{\textit{SBCE loss}}. Development Set (top three rows), De-duplicated Test Set (middle three rows), Original Test Set (bottom three rows). \textbf{*} denotes statistically significant (p < 0.05) from baseline. Bold figures indicate the best performing method(s) for each of the two metrics and for each row (experimental setting).}
\label{soft_loss_table}
\end{table*}

\subsection{BCE and SBCE Loss}

Conventional BCE loss encourages the policy to learn all available actions, aiming to maximise the HardF1 score on the test set. Training with SBCE may lead to less diverse agent responses, however, both intrinsic and extrinsic scores show consistent improvements over BCE of around 5 SoftF1 points for M2M datasets and 10 points for MultiWOZ. The declining HardF1 scores \textit{do not} correlate with lower success rates in end-to-end evaluation (see Table~\ref{convlab_table}). The strongest effect of SBCE is that most differences between experiments observed in Table \ref{hard_loss_table} were neutralised. HardF1 scores decrease to roughly same levels while SoftF1 scores increase to their highest levels (with a few statistically significant results). We think this is because the SBCE loss may lead the policy to converge to approximately the same actions. We also observe that even without augmentation, the DM policy can be significantly improved with SBCE alone.

\subsection{Downsampling}

Downsampling reduces the original data size by filtering out duplicate instances. This has a similar effect as oversampling infrequent examples, i.e. reducing biases towards some training instances. Downsampling ignores the likelihood of agent actions as observed in the original data, an effect also seen with uniform sampling in Section~\ref{data_augmentation}. For M2M restaurant, which has the highest repetition and therefore the most biased paths through the graph, this leads to a substantial decline in F1-scores. Specifically: i) the 'rating' slot score dropped by 70\% ii) the 'time' and 'date' slots dropped by 41\% and iii) the 'confirm' intent decreased by 26\%. Similar trends were observed in MultiWOZ scores but to a lesser extent. The M2M movie dataset contains relatively little repetition hence we observed no significant changes. Downsampling is more suitable for dialogues with more evenly distributed agent actions.

\subsection{Data Duplication}

Training data duplication did not produce any significant changes as compared to the baseline. Over all dialogue histories, the SoftF1 and HardF1 scores fluctuate around the original training data scores without any consistent patterns. This augmentation only seems effective in ultra-low data regimes \cite{bocklisch2017rasa} where one possesses at most a few dozen training dialogues.

\begin{table*}[h]
\centering
\begin{tabular}{l||r|r||r|r||r|r||r|r}
\toprule
\multicolumn{1}{c||}{} & \multicolumn{2}{c||}{BASE (B)} & \multicolumn{2}{c||}{MFS} & \multicolumn{2}{c||}{B+MFS} & \multicolumn{2}{c}{Oracle} \\ 
\hline
 & Success & \#Turns & Success & \#Turns & Success & \#Turns & Success & \#Turns\\ \hline
BCE  & 73.4\% & 10.11 & 72.4\% & 9.91 & \textbf{79.8}\% & 9.03 & 70.2\% & 10.43\\ \hline
SBCE & 75.3\% & 9.01  & \textbf{76.0}\% & 9.60 & \textbf{76.0}\% & 9.55 & 73.8\% & 9.24\\ 
\bottomrule
\end{tabular}
\caption{MultiWOZ success rate and the average number of turns with a user simulator over 1000 dialogues. }
\label{convlab_table}
\end{table*}

\subsection{Most Frequent Sampling} \label{mfs}

MFS generates novel training instances so that the most frequent agent actions are preceded by new histories, that is, one or more original paths leading to common actions. Due to this, the infrequently visited edges effectively get pruned from the graph, leading to a $\sim$20\% reduction in size compared to baseline data for MultiWOZ, $\sim$50\% for M2M (M) and $\sim$60\% reduction for M2M (R). Repetition is also removed so that each training example occurs exactly once. As a consequence of that, the overlap between MFS train and dev/test sets is also reduced by 40\%-60\% compared to baseline. In spite of having substantially fewer paths through the graph, this is the most effective intervention from a data standpoint. We should note that when combined with SBCE, MFS is no longer considering the most probable action exclusively as the training will defer to an equally valid action if the calculated loss is lower. Due to this, MFS achieves the highest SoftF1 with BCE loss without sacrificing HardF1. In cases where MFS alone is less effective, it can be combined with the baseline training data to achieve best performance. MFS achieves the best HardF1 when combined with SBCE (except M2M restaurant). In extrinsic evaluation, after 1000 simulated conversations with a user, MFS combined with the baseline train data improves success rate from 73.4\% to 79.8\% with BCE loss. Success rate also improves with SBCE loss to 76\% with and without adding original data. The length of the dialogue is also consistently reduced as the agent satisfies the user's goals faster.\\

We observe no distinct error patterns for the movie task except for the lack of usage of the 'offer' intent, even with MFS data. This may be due to the lower frequency of the 'offer' intent in the dataset relative to other dialogue acts. Instead, we observe consistent, single digit improvements ($\sim$4 F1 points) for almost all actions and slots. For M2M restaurant, there are two main patterns: i) the most problematic errors discussed in the downsampling section were reversed. It now means that the 'rating', 'time', 'date' and 'confirm' targets show a good improvement rather than a decline. Even the previously unused 'meal' slot went from 0 to 35 points. ii) the remaining actions and slots show a single digit improvement similar to the movie task. The M2M restaurant is particularly sensitive to the removal of repeated instances hence benefits from additional training on frequent agent actions. Errors in MultiWOZ were also reduced although two domains (police and hospital) have not been learnt despite the additional MFS data. This is likely owing to their very low frequency in the training data. For all other domains, we observe that an estimated one third of the dialogue acts that were unused with baseline training data advance by around 20 F1 points, on average. More frequent dialogue acts have also improved by an estimated 10 F1 points. Despite the advantages of augmentation, many dialogue acts are still predicted with low accuracy (or not at all), which explains where the remaining $\sim$20\% success score in extrinsic evaluation and $\sim$26 F1 points in intrinsic evaluation could be recovered. Example dialogues are provided in the Appendix.

\subsection{Oracle Augmentation}

Oracle generates novel instances at $\sim$56\% of the original train data size for M2M restaurant, $\sim$65\% for M2M movie but only $\sim$3\% for MultiWOZ due to the small percentage of shared edges. As expected, we observed improvements over baseline, however, with some caveats. Oracle augmentation illustrates the need for the usage of both soft and hard evaluation. Using the conventional approach, BCE training with HardF1 evaluation, we would correctly conclude that Oracle is (mostly) effective for M2M datasets and only marginally so for MultiWOZ. While SoftF1 scores also improve over the baseline, those are 1-2 points lower than the best MFS scores. For MultiWOZ, this difference is even greater (5-6 points lower) and would be indiscernible with only a HardF1 score available to guide experimentation. Oracle augmentation is an example where the policy training risks overfitting to maximise HardF1 at the expense of SoftF1, which may explain the 2-3\% decline in success rate in end-to-end evaluation. When oracle augmentation is effective (movie task), the error pattern is similar to the movie MFS experiment. In other words, more accurate predictions were observed for all dialogue acts but with a lower magnitude.

\section{Future Work}
\label{future}

\subsection{Transformers}
Our data augmentation method is orthogonal to the choice of the (sequence) model. We have used an LSTM for all experiments. We have also briefly tested a Multi Layer Perceptron where the input consisted of concatenated time steps, yielding comparable results to the LSTM model. Other architectures such as Transformers \cite{vaswani2017attention}, which have recently achieved SOTA performance on language modelling and transfer learning, can also be used. However, due to the \textit{symbolic} nature of the dialogue management input, we may not see an advantage from using pretrained transformers that compute representations of \textit{natural language}. Also, as we previously mentioned (see Section~\ref{experimental}), we have performed experiments with LSTMs using larger hidden state sizes but they did not lead to any improvement. 
We don't expect any significant improvement in performance by switching the architecture to Transformers.

\subsection{Semi-Random Data Augmentation}

A random graph traversal augmentation should be avoided as the dialogue flows in the train, development and test sets (including the user simulator) are not random. Some paths are more likely than others and some nodes/edges are more frequently visited than others. A more promising approach to show policy improvement may be with semi-random sampling from the train ConvGraph, using the validation set performance as a reward signal. Similar to a hyperparameter search (even reinforcement learning), one can repeatedly sample training instances from different hyperparameters until a stopping criterion is met. Though more computationally intensive and more challenging to reproduce, this type of data augmentation may deliver novel insights into the generation process.

\subsection{Data Generation with ConvGraphs}
\label{data_gen}

ConvGraph is expected to be initialised from existing dialogues in order to augment the training data. However, we can also collect new data with ConvGraph by \textit{checking the uniqueness} of incoming dialogue turns, possibly in real-time. New nodes and edges will make the graph denser and allow for maximally diverse data augmentation, avoiding needless repetition and accelerating data collection. ConvGraph can be efficiently expanded as the environment or user behaviours change over time in order to extend an artificial agent with additional capabilities or to bootstrap agent policies interactively \cite{williams2017demonstration,williams2017hybrid,liu2018dialogue}. 

\subsection{RNN ConvGraph}
We also propose the RNN-ConvGraph, a theoretical alternative to ConvGraph. This is a generative model that takes as input a sequence of previous dialogue states $PDS$ = $[ds_{t-1},\ ds_{t-2},\ ...,\ ds_{t-n}]$ where $n$=$max\_history$ and predicts the next dialogue state $ds_{t}$ over a 'vocabulary' of all graph nodes. Reminiscent of a generative language model, RNN-ConvGraph augments training data by using the conditional probabilities learned from the train data. Instead of an explicit graph data structure, the RNN-ConvGraph would be an implicit representation of the graph. 

\section{Conclusions}

We have introduced the Conversation Graph for Dialogue Management, an approach that unifies conversations based on matching nodes (dialogue states). Exploiting the structure of ConvGraph can be effectively used for 1) data augmentation with our Most Frequent Sampling method, 2) training non-deterministic policies with SBCE, our soft loss function and 3) a more complete and fair evaluation of non-deterministic agents with our SoftF1 score. We conducted a thorough analysis of ConvGraph on three conversational datasets and showed that they can have markedly different properties. Extrinsic evaluation with a user simulator as well as intrinsic evaluation supports that ConvGraph can successfully augment datasets by generating novel paths through the graph. The soft training loss SBCE lets the agent choose which actions to learn in each dialogue state, leading to consistent policy improvements. Finally, the soft evaluation has extended the conventional 'hard' evaluation, which was insufficient for non-deterministic agents, leading to unfair penalties for correct predictions. We hope that our methodology as well as suggestions for future work will inspire further research in this topic area.

\section*{Acknowledgements}
We would like to thank the reviewers for their thoughtful and insightful comments, which we found very helpful in improving the paper.

\bibliography{tacl2018}
\bibliographystyle{acl_natbib}

\clearpage
\appendix
\section{Dialogue Examples - MultiWOZ}

The following tables show examples of typical errors committed on the MultiWOZ dataset. For clarity and brevity, we do not provide the dialogues in full, only the turns relevant to the models' decisions. We set the dialogue history length to $n$=3 thus $PDS$=[$ds_{t-1}$, $ds_{t-2}$, $ds_{t-3}$] (see Section~\ref{data_augmentation}). At turn $t$, we show output for models BASE and B+MFS (see Section~\ref{mfs}). For more detailed analysis of each model's strengths and failings, we refer the reader back to Section~\ref{results}. \\

\begin{table}[ht]
\begin{center}
\resizebox{\columnwidth}{!}{
\begin{tabular}{m{1cm}| m{4cm} m{4cm}}
\textbf{Turn} & \textbf{User} & \textbf{Agent} \\ \hline
t-3 & Hotel-Inform(Parking) & \\
t-2 & & Hotel-Inform(Choice), Hotel-Request(Price)\\
t-1 & Hotel-Inform(Price), & \\
 & Hotel-Request(Area),  & \\
 & Restaurant-Inform(Price) & \\ \hline \hline
t & & \textbf{B+MFS:}  \\
& & Booking-Inform(None) \\
& & \textbf{BASE (B):} \\
& & Booking-Inform(None), \\
& & Hotel-Inform(Name) \\
\end{tabular}}
\end{center}
\caption{MultiWOZ example \#1}
\label{multiwoz_example_1}
\end{table}

In Table~\ref{multiwoz_example_1}, at turn $t-3$, the user tells the agent of that he's interested in free hotel parking \textit{Hotel-Inform(Parking)}. The agent replies that there are many available hotels the user can choose from \textit{Hotel-Inform(Choice)}. The agent additionally asks the user to provide a desirable price range to further filter down the choices \textit{Hotel-Request(Price)}. At the next turn $t-1$, the user provides a specific price-range, asks about the hotels' area and requests that the restaurant they are booking in parallel should be in a specific price-range. We remind the reader that in MultiWOZ, the same dialogue may span multiple domains. In the current turn $t$, both models respond that no hotel could be found with the additional criteria \textit{Booking-Inform(None)}. Inappropriately, however, the BASE model also provides a hotel's name \textit{Hotel-Inform(Name)} but that hotel would not be meeting the criteria.\\

\begin{table}[ht]
\begin{center}
\resizebox{\columnwidth}{!}{
\begin{tabular}{m{1cm}| m{4cm} m{4cm}}
\textbf{Turn} & \textbf{User} & \textbf{Agent} \\ \hline
t-3 & Hotel-Inform(Parking) & \\
t-2 & & Hotel-Inform(Choice)\\
t-1 & Hotel-Inform(Price) & \\ \hline \hline
t & & \textbf{B+MFS:}  \\
& & Booking-Inform(None), \\
& & Hotel-Recom(Area), \\
& & Hotel-Recom(Name) \\
& & \textbf{BASE (B):} \\
& & Booking-Inform(None) \\
\end{tabular}}
\end{center}
\caption{MultiWOZ example \#2}
\label{multiwoz_example_2}
\end{table}

In Table~\ref{multiwoz_example_2}, we observe a similar history that leads both agents to declare that no hotels could be found under the criteria. However, the B+MFS model goes a step further and recommends a hotel in a different area \textit{Hotel-Recom(Name, Area)}. In extrinsic evaluation, such action will likely lead to a shorter dialogue (see \#turns in Table \ref{convlab_table}). Proactively offering appropriate hotel choices should also lead to a higher dialogue success rate.\\

\begin{table}[ht]
\begin{center}
\resizebox{\columnwidth}{!}{
\begin{tabular}{m{1cm}| m{4cm} m{4cm}}
\textbf{Turn} & \textbf{User} & \textbf{Agent} \\ \hline
t-3 & Hotel-Request(Phone) & \\
t-2 & & Hotel-Inform(Phone)\\
t-1 & Train-Inform(Dest) & \\ \hline \hline
t & & \textbf{B+MFS:}  \\
& & Train-Inform(Choice) \\
& & Train-Request(Leave) \\
& & \textbf{BASE (B):} \\
& & Train-Request(Leave) \\
\end{tabular}}
\end{center}
\caption{MultiWOZ example \#3}
\label{multiwoz_example_3}
\end{table}

In Table~\ref{multiwoz_example_3}, both models request additional information from the user \textit{Train-Request(Leave)}. However, the B+MFS agent also informs the user that there are many choices available and offers to book the train. The inclusion of a 'call to action' increases user engagement and the probability of a successful train booking.\\

\begin{table}[ht]
\begin{center}
\resizebox{\columnwidth}{!}{
\begin{tabular}{m{1cm}| m{4cm} m{4cm}}
\textbf{Turn} & \textbf{User} & \textbf{Agent} \\ \hline
t-3 & Train-Inform(Dest) & \\
t-2 & & Train-Request(Leave)\\
t-1 & Train-Inform(Leave) & \\ \hline \hline
t & & \textbf{B+MFS:}  \\
& & Train-Inform(Arrive) \\
& & Train-Inform(ID) \\
& & Train-OfferBook() \\
& & \textbf{BASE (B):} \\
& & Train-Inform(ID) \\
\end{tabular}}
\end{center}
\caption{MultiWOZ example \#4}
\label{multiwoz_example_4}
\end{table}

The final example for MultiWOZ in Table~\ref{multiwoz_example_4} shows that after collecting all required information, the BASE agent merely informs the user that it found a train (specified by its ID) that meets the criteria. The B+MFS agent additionally informs the user of the train's arrival time and offers to book the train for the user.

\clearpage
\section{Dialogue Examples - Machines to machines (M2M)}

The following examples are taken from the M2M dataset (the restaurant domain) in order to illustrate some of the typical errors of the baseline (BASE) and our best augmented method (B+MFS). More error analysis is provided in Section~\ref{results}.\\

\begin{table}[ht]
\begin{center}
\resizebox{\columnwidth}{!}{
\begin{tabular}{m{1cm}| m{4cm} m{4cm}}
\textbf{Turn} & \textbf{User} & \textbf{Agent} \\ \hline
t-3 & Affirm() & \\
t-2 & & Request(Date)\\
    & & Request(Time)\\
t-1 & Inform(Date) & \\ 
    & Inform(Time) & \\ \hline \hline
t & & \textbf{B+MFS:}  \\
& & Confirm(Time) \\
& & \textbf{BASE (B):} \\
& & Request(Time) \\
\end{tabular}}
\end{center}
\caption{M2M (R) example \#1}
\label{m2m_example_1}
\end{table}

In Table~\ref{m2m_example_1}, at turn $t-1$, the user provides a time \textit{Inform(Time)} and date \textit{Inform(Date)} for the restaurant booking. The BASE model asks for the time again, which is clearly problematic as it seems to ignore the user's input. The B+MFS agent correctly handles the query and asks the user to confirm that choice thus avoiding a penalty.\\

\begin{table}[ht]
\begin{center}
\resizebox{\columnwidth}{!}{
\begin{tabular}{m{1cm}| m{4cm} m{4cm}}
\textbf{Turn} & \textbf{User} & \textbf{Agent} \\ \hline
t-3 & Inform(Category) & \\
t-2 & & Confirm(Category)\\
t-1 & Affirm() & \\ \hline \hline
t & & \textbf{B+MFS:}  \\
& & Request(Meal) \\
& & \textbf{BASE (B):} \\
& & Select(Restaurant\_Name) \\
& & Select(Restaurant\_Name) \\
& & Select(Restaurant\_Name) \\
\end{tabular}}
\end{center}
\caption{M2M (R) example \#2}
\label{m2m_example_2}
\end{table}

In Table~\ref{m2m_example_2}, the user lets the agent know about the preferred type of restaurant \textit{Inform(Category)} they would like to book. When this is confirmed by the user, the BASE model is prematurely asking the user to select from a list of available restaurants before all necessary slots have been filled. The B+MFS agent correctly requests that the user specifies what type of meal they are interested in first.\\

\begin{table}[ht]
\begin{center}
\resizebox{\columnwidth}{!}{
\begin{tabular}{m{1cm}| m{4cm} m{4cm}}
\textbf{Turn} & \textbf{User} & \textbf{Agent} \\ \hline
t-3 & Find\_Restaurant() & \\
t-2 & & Request(Category)\\
t-1 & Inform(Category) & \\ \hline \hline
t & & \textbf{B+MFS:}  \\
& & Request(Location) \\
& & Request(Rating) \\
& & \textbf{BASE (B):} \\
& & Request() \\
\end{tabular}}
\end{center}
\caption{M2M (R) example \#3}
\label{m2m_example_3}
\end{table}

Table~\ref{m2m_example_3} shows the user asking for a specific category of restaurant \textit{Inform(Category)}. The BASE model only provides a partial output, correctly predicting that it should request information from the user but not which information it requires. This means the user would have to repeat the query in the better case or lose a dialogue in the worst case. \\

\begin{table}[ht]
\begin{center}
\resizebox{\columnwidth}{!}{
\begin{tabular}{m{1cm}| m{4cm} m{4cm}}
\textbf{Turn} & \textbf{User} & \textbf{Agent} \\ \hline
t-3 & Inform(Price\_Range) & \\
t-2 & & Confirm(Price\_Range)\\
t-1 & Affirm() & \\ \hline \hline
t & & \textbf{B+MFS:}  \\
& & Select(Restaurant\_Name) \\
& & Select(Restaurant\_Name) \\
& & Select(Restaurant\_Name) \\
& & \textbf{BASE (B):} \\
& & Offer(Restaurant\_Name) \\
\end{tabular}}
\end{center}
\caption{M2M (R) example \#4}
\label{m2m_example_4}
\end{table}

Table~\ref{m2m_example_4} shows the user asking for a restaurant in a specific price range \textit{Inform(Price\_Range)}. Once the agent confirms that this is what the user is looking for, the BASE agent proceeds to offer a single restaurant. While this is not wrong, the B+MFS model offers three choices using the \textit{Select(Restaurant\_Name)} dialogue act. A preference for this action may lead to a higher user satisfaction and ultimately to more successful dialogues.

\end{document}